# Robust Report Level Cluster-to-Track Fusion

Johan Schubert
Department of Data and Information Fusion
Division of Command and Control Systems
Swedish Defence Research Agency
SE–172 90  Stockholm, Sweden
`schubert@foi.se`
`http://www.foi.se/fusion/`

**Abstract** - *In this paper we develop a method for report level tracking based on Dempster-Shafer clustering using Potts spin neural networks where clusters of incoming reports are gradually fused into existing tracks, one cluster for each track. Incoming reports are put into a cluster and continuous reclustering of older reports is made in order to obtain maximum association fit within the cluster and towards the track. Over time, the oldest reports of the cluster leave the cluster for the fixed track at the same rate as new incoming reports are put into it. Fusing reports to existing tracks in this fashion allows us to take account of both existing tracks and the probable future of each track, as represented by younger reports within the corresponding cluster. This gives us a robust report-to-track association. Compared to clustering of all available reports this approach is computationally faster and has a better report-to-track association than simple step-by-step association.*

## 1 Introduction

In this paper we develop a robust incremental cluster-to-track fusion algorithm with an ability to take a second look on already made report-to-track associations. This is achieved through using Dempster-Shafer clustering (DSC) and Potts spin (PS) neural networks in an incremental manner (iPDSC) where incoming reports are gradually fused into existing tracks.
DEFINITION. Dempster-Shafer clustering is any method of clustering uncertain data using the conflict in Dempster's rule (or a function thereof) as distance measure.

We use a logarithmization of conflicts in Dempster-Shafer theory [1] in order to map DSC onto a PS neural network.

The cluster-to-track fusion described is an extension of earlier clustering methods for clustering large scale problems in a non-incremental way [2–6]. This methodology has been further extended [7] for clustering reports in a non-incremental way based on both attracting (e.g., communications patterns) and conflicting information (i.e., inter-intelligence inconsistency), although only conflicting informations is used here.

Similar applications in information fusion on correlation problems using the conflict between belief function are found in [8–12] and using possibility theory in [13].

In Section 2 we describe the memory management used for dynamic clustering. The Potts spin Dempster-Shafer clustering process is then described in Section 3 for the case with short- and long-term memory. Finally, a series of test runs are performed to evaluate clustering performance and computation time for different sizes of short- and long-term memory.

## 2 Memory management

Previously we used an intelligence management process where a high number of intelligence reports were clustered at each opportunity [11]. If new intelligence arrived during an ongoing clustering they could immediately be classified by comparison with the previous cluster result of older intelligence, but it was impossible to constantly recluster newly arrived intelligence with older intelligence due to time complexity. We had to wait until an ongoing clustering process terminated before we could start a new clustering with all new intelligence that had arrived during the latest clustering process.

Here, we develop a dynamic clustering process where reclustering is made continuously when new intelligence arrive at the process. We have organized all current intelligence in three different memories. First, the most recently arrived intelligence is put into a short-term memory. Secondly, the older intelligence is maintained in a long-term memory, and finally, intelligence no longer used in clustering, is kept in a history memory used only for evaluation of clustering performance.

When a piece of new intelligence arrives at the system it is immediately put into the short-term memory, while at the same time the oldest intelligence report in the short-term memory is moved to the long-term memory where intelligence-to-cluster associations are maintained for the future. Similarly, the oldest report in the long-term memory is moved to the history where it is kept for further investigation and for evaluation of clustering performance purposes, Figure 1.

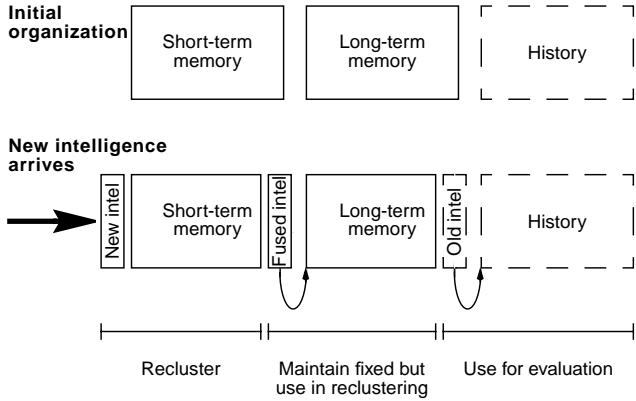

Figure 1. Memory management.

A complete reclustering of all intelligence in the short-term memory is made using the Potts spin Dempster-Shafer clustering process (PDSC). In clustering we take into account the conflicts between all intelligence reports of both the short-term and long-term memories while only updating the cluster association of the intelligence in the short-term memory. Thus, all intelligence in long-term memory remains fixed during clustering of short-term memory, with intelligence report-to-cluster association unchanged for all time.

After clustering of short-term memory, short-term and long-term memories together with the history jointly make up the current view of tracks and all report to track association. Long-term memory and history together make up a permanent view on report-to-track association that is not put into question by new intelligence, while the report-to-track association of short-term memory may change if new intelligence becomes available.

## 3 Clustering of short-term memory

We perform clustering of all reports in short-term memory using Dempster-Shafer clustering [5] and Potts spin [14] mean field theory. With this method we minimize a distance measure of weight of conflict by neural clustering. A summary of the Potts spin method and its application to information fusion and intelligence analysis was presented in [11]. We will in this section provide a short sketch of the method for the sake of completeness.

We use the clustering process to separate the intelligence into subsets for the tracked objects. We combine Dempster-Shafer theory with the Potts spin neural network model into a powerful solver for large scale Dempster-Shafer clustering problems.

In [15] we established a criterion function of overall conflict called the metaconflict function. The metaconflict is derived as the plausibility that the partitioning of all reports is correct when the conflict in each subset is viewed as evidence against the partition.

DEFINITION. *Let the* metaconflict function,

$$Mcf(K, S_1, S_2, ..., S_n) \triangleq 1 - \prod_{i=1}^{K}(1-c_i), \quad (1)$$

*be the conflict against a partitioning of n reports of the set* $\chi$ *into K disjoint subsets* $\chi_i$. *Here,* $c_i$ *is the conflict in subset i.*

We use the minimization of the metaconflict as the method of partitioning the reports into subsets; each subset referring to a separate track to be treated independently.

The Potts model handle data as pairwise terms. It is therefore necessary to simplify this conflict function, and write the conflict as a sum of pairwise conflicts, i.e., to linearize the conflict function by taking its logarithm. We rewrite [5] the minimization as follows

$$\min Mcf$$
$$\Leftrightarrow$$
$$\max \log(1 - Mcf) = \max \log \prod_i (1 - c_i)$$
$$= \max \sum_i \log(1 - c_i) = \min \sum_i -\log(1 - c_i) \quad (2)$$
$$\approx \sum_i \sum_{\substack{k,l \\ S_k, S_l \in \chi_i}} -\log(1 - s_k s_l).$$

The Potts spin problem consists of minimizing an energy function

$$E = \frac{1}{2} \sum_{i,j=1}^{N} \sum_{a=1}^{q} J_{ij} S_{ia} S_{ja} \quad (3)$$

by changing the states of the $S_{ia}$'s, where $S_{ia} \in \{0, 1\}$ and $S_{ia} = 1$ means that report $i$ is in cluster $a$. This model serve as a clustering method if $J_{ij}$ is used as a penalty factor when report $i$ and $j$ are in the same cluster; reports in different clusters get no penalty.

In Dempster-Shafer clustering we use the conflict of Dempster's rule when all elements within a subset are combined as an indication of whether these reports belong together. The higher the conflict is, the less credible that they belong together.

In order to apply the Potts model to Dempster-Shafer clustering we use $J_{ij} = -\log(1-s_i s_j)\delta_{|A_i \cap A_j|}$, where $s_i$ and $s_j$ are basic probability numbers of reports $i$ and $j$. By minimizing the energy function we also minimize the overall conflict.

The minimization is carried out by simulated annealing. In simulated annealing temperature is an important parameter. The process starts at a high temperature where the $S_{ia}$ change state more or less at random and begin to lower the temperature gradually. As the temperature is lowered the spins become biased by the interactions ($J_{ij}$) reaching a minimum of the energy function. This also gives us the best partition of all intelligence into the clusters with minimal overall conflict.

For computational reasons we use a mean field model, where spins are deterministic with $V_{ia} \in [0, 1]$, in order to find the minimum of the energy function. The Potts mean field equations are derived [16] as

$$V_{ia} = \frac{e^{-H_{ia}[V]/T}}{\sum_{b=1}^{K} e^{-H_{ib}[V]/T}} \quad (4)$$

where

$$H_{ia}[V] = \frac{\sum_{j=1}^{N} J_{ij} V_{ja} - \gamma V_{ia} + \alpha \sum_{j=1}^{N} V_{ja}}{\frac{K}{N} \sum_{i=1}^{N} V_{ia}}. \quad (5)$$

In order to minimize the energy function Eqs. (4) and (5) are used recursively for all $i$ that belong to short-term memory until a stationary equilibrium state has been reached for each temperature. Then, the temperature is lowered step by step by a constant factor until $V_{ia} = 0, 1 \quad \forall i, a$ in the stationary equilibrium state, Figure 2.

## 4 Experiment

We have performed an experiment where we receive 125 reports about seven different targets. Our frame of discernment, i.e., the targets, are $\Theta = \{A, B, C, D, E, F, G\}$ and we have 127 possible different reports about the targets $2^\Theta - \emptyset$, i.e., all subsets of the frame.

We create ten different data sets where the 125 reports in each data set are drawn in a different random order without replacement from the set of 127 reports and are given a random basic probability number uniformly between [0, 1] to represent the uncertainty of the report.

We run tests for different sizes of short-term and long-term memory. The short-term memory varies between five and 100 in size with an increment of five. The long-term memory varies between zero and 100 minus the size of short-term memory. Thus, the sum of short-term and long-term memories are never greater than 100. In total we perform 210 test with different sizes of short-term and long-term memories. For each test we start with a number of reports equal to the size of short-term memory and cluster these. We then receive further reports one-by-one and recluster short-term memory after each new report. At each step the oldest report in the short-term memory is moved to the long-term memory. This goes on until a total of 100 reports are drawn. Both memories are now filled up and any additional reports have been transferred to the history, e.g, if short-term memory has a size of 20, with long-term memory 30 in size, we now have 20 and 30 reports respectively in those memories and 50 reports have been transferred to the history.

**INITIALIZE**

$K$ (No. of clusters); $N_l$ (No. of elements in long-term memory); $N_s$ (No. of elements in short-term memory); $N = N_l + N_s$; $S_i$ (Report no. $i$);
$J_{ij} = -\log(1 - s_i s_j) \delta_{|A_i \cap A_j|} \quad \forall 1 \leq i, j \leq N$;
$s = 0$; $t = 0$; $\varepsilon = 0.001$; $\tau = 0.9$; $\alpha$ (for $K \leq 7$: $\alpha = 0$, $K = 8$: $\alpha = 10^{-6}$, $K = 9$: $\alpha = 0$, $K = 10$: $\alpha = 3 \cdot 10^{-7}$, $K = 11$: $\alpha = 3 \cdot 10^{-8}$); $\gamma = 0.5$;
$T^0 = T_c$ (a critical temperature) $= \frac{1}{K} \cdot max(-\lambda_{min}, \lambda_{max})$, where $\lambda_{min}$ and $\lambda_{max}$ are the extreme eigenvalues of M, where $M_{ij} = J_{ij} + \alpha - \gamma \delta_{ij}$;

$$V_{ia}^0 = \begin{cases} 1, & \forall 1 \leq i \leq N_l, S_i \in \chi_a \\ 0, & \forall 1 \leq i \leq N_l, S_i \notin \chi_a \\ \frac{1}{K} + \varepsilon \cdot rand[0,1], & \forall N_l < i \leq N, a \end{cases}$$

**REPEAT**
- REPEAT-2
  - $\forall a \quad G_a^s = \frac{K}{N} \sum_{i=1}^{N} V_{ia}$;
  - $\forall N_l < i \leq N$ Do:
  - $H_{ia}^s = \frac{\sum_{j=1}^{N} (J_{ij} + \alpha) V_{ja}^s + J_{i(N+a)} + \alpha - \gamma V_{ia}^s}{G_a^s} \quad \forall a$;
  - $F_i^s = \sum_{a=1}^{K} e^{-H_{ia}^s/T^t}$;
  - $V_{ia}^{s+1} = \frac{e^{-H_{ia}^s/T^t}}{F_i^s} + \varepsilon \cdot rand[0,1] \quad \forall a$;
  - $s = s + 1$;
  UNTIL-2
  $\frac{1}{N} \sum_{i,a} (V_{ia}^s - V_{ia}^{s-1}) \leq 0.01$;
- $T^{t+1} = \tau \cdot T^t$;
- $t = t + 1$;

**UNTIL**
$\frac{1}{N} \sum_{i,a} (V_{ia}^s)^2 \geq 0.99$;

**RETURN**
$\{\chi_a | \forall S_i \in \chi_a, \forall b \neq a \quad V_{ia}^s > V_{ib}^s\}$;

Figure 2. Clustering.

For the sake of the clustering algorithm we could have been content with the first 50 reports for filling up the short-term memory and successively building up the long-term memory, but in order to measure the clustering performance in a consistent manner we need to have an equal number of reports in the short-term memory, long-term memory and history put together.

Finally, we start receiving the last 25 reports one-by-one with reclustering of short-term memory after each report. After each of the 25 clustering processes the total weight of conflict among the latest 100 reports are summed up and the average of this measure is calculated after having received the last 25 reports.

This entire process is then repeated for all ten data sets and the average weight of conflict for the ten data sets is calculated.

In total we perform 187 600 clustering processes for all the configurations and data sets, out of which 135 100,

$$\left\{ \sum_{i=1}^{20} i \cdot [1 + 5 \cdot (i-1)] \right\} \times 10, \quad (6)$$

are used for building up long-term memory and history for the first 100 reports. In order to measure the performance for each configuration and data set, we then perform 52 500 (210x25x10) clustering processes for the 210 different sizes of short-term and long-term memories, 250 clustering processes per configuration (25x10).

## 5 Results

In Figure 3 we have plotted the average classification error rate for all 210 different configurations of short-term and long-term memory. We notice in Figure 3 that roughly speaking we obtain a good classification rate when the joint size of short-term and long-term memory is maximal, i.e., equal to 100. We can here obtain an error rate below 0.5%.

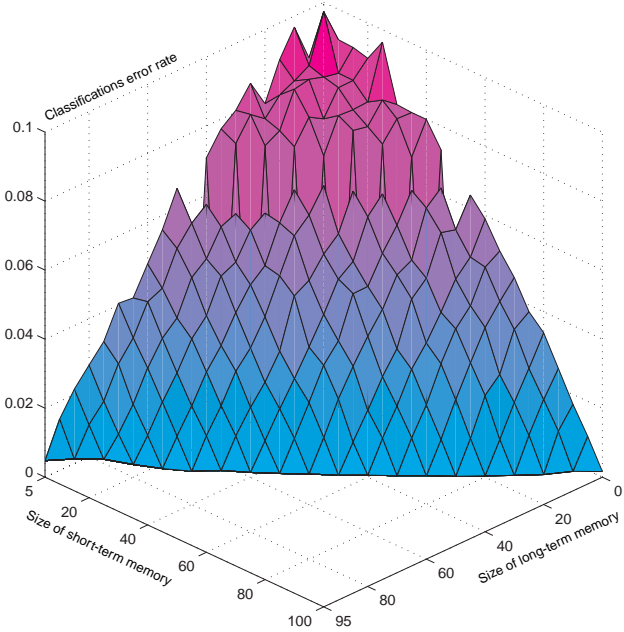

Figure 3. Classification error rate for different sizes of short-term and long-term memories.

The classification error rate is calculated by comparing the total weight of conflict per cluster with the expected weight of conflict from one misclassified report.

First, the total weight of conflict is calculated from summation of the pairwise weights of conflict between all 100 reports whether they are in the short-term memory, the long-term memory or in the history.

Secondly, with $2^K - 1$ different reports, all simple support functions with elements from the set of all subsets $2^\Theta$ of $\Theta = \{A, B, C, D, E, F, G\}$, there are

$$\frac{1}{2}[(2^K - 1)^2 - (2^K - 1)] \quad (7)$$

possible pairs of two pieces of evidence, e.g., 8001 pairs when $K = 7$. Of these,

$$\frac{1}{2} \sum_{j=1}^{K-1} \binom{K}{j} \sum_{i=1}^{K-j} \binom{K-j}{i} \quad (8)$$

are in conflict, e.g., 966 pairs when $K = 7$.

- Thus, if two different pieces of evidence are drawn randomly from the set of all subsets, we have a probability of conflict between their propositions of

$$P(A_i \cap A_j = \varnothing) = \frac{\sum_{j=1}^{K-1} \binom{K}{j} \sum_{i=1}^{K-j} \binom{K-j}{i}}{(2^K - 1)^2 - (2^K - 1)}, \quad (9)$$

where, e.g., $P(A_i \cap A_j = \varnothing) = 12.1\%$ when $K = 7$.

- The basic probability number of a report in our test is a uniformly distributed random number between 0 and 1. Thus, the expected conflict is 0.25 between two pieces of evidence that are known to be in conflict, i.e., an expected weight of conflict of $-\log(1 - 0.25) = 0.288$. Therefore, the expected weight of conflict between two pieces of evidence drawn randomly from $2^\Theta$ becomes $0.288 \cdot 0.121 = 0.0347$.

- As we have $100/K$ (= 14.3) reports on average per cluster in this performance evaluation, we have on average 94.9 pairs per cluster.

This gives us an expected weight of conflict per misclassification of $0.0347 \cdot 94.9 = 3.30$.

Thirdly, the weight of conflict per cluster may now be divided by the weight of conflict per misclassification (= 3.30) to yield the number of misclassifications per target, with the result further divided by the average number of reports per cluster (= 14.3) to yield the classification error rate.

We should note that while a high conflict between two reports in the same cluster is proof of misclassifications, no conflict is not proof of a perfect classification due to the allowed representation for reports in Dempster-Shafer theory. Reports may refer to a set of targets, e.g., two reports with propositions $\{A, B\}$ and $\{B, C\}$ are never interpreted as misclassified when they appear in the same cluster, regardless of which cluster that is, since they have a nonempty intersection with zero conflict. Thus, the error rates may not be directly comparable to error rates in Bayesian theory where sensors must only report on single targets. The error rates in this paper are used to compare performance of different configurations of short-term and long-term memories.

In Figure 4 we have plotted the average computation time for the same 210 configurations. We notice that the computation time is mostly dependent on the size of the short-term memory. For fairly small short-term memories we receive a computation time of about one second.

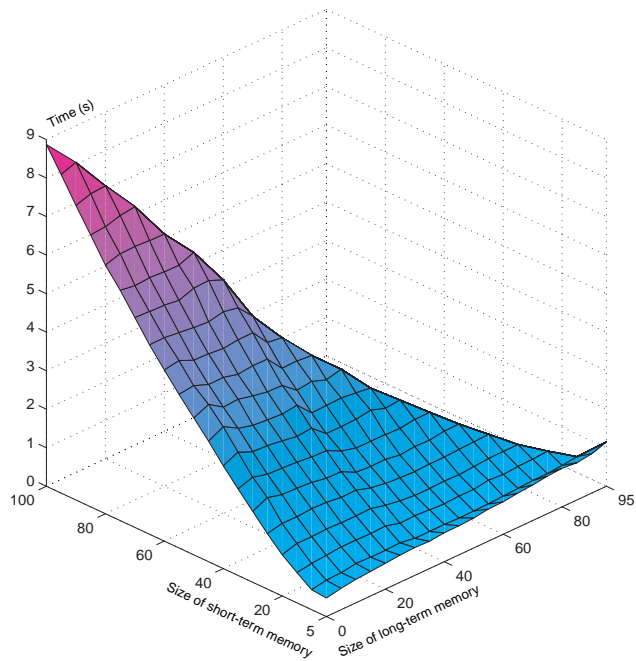

Figure 4. Computation time is seconds.

A good configuration would then be a fairly small short-term memory with a large long-term memory in order to obtain close to maximal clustering performance and a decent computation time. A closer look is made in Figure 5 where the classification error rate and average computation time is plotted for three joint sizes of short-term and long-term memories for different sizes of short-term memory.

Again we observe that it is crucial for clustering performance to have a maximal joint size of short-term and long-term memory, especially since the computation time is virtually unchanged when the size of long-term memories decreased for a fixed size of short-term memory. This is much more important than the actual sizes of the two memories.

While a choice of actual configuration is always domain dependent it is obvious that in this test example good performances can be found with sizes of short-term memory between 20 and 50 as classification errors are below 0.5% and computation time is reasonable for many problems.

For short-term memories with sizes of 20, 30, 40 and 50 (with a joint size of 100) we have classification error rates of 0.36%, 0.17%, 0.14% and 0.09%, respectively, and computation times of 1.1, 1.6, 2.2 and 3.0 seconds, respectively, Table 1. The lowest classification error rate was 0.024% with a computation time of 6.1 seconds (for a short-term memory of size 75) and the fastest computation time was 0.77 seconds with a classification error rate of 0.53% (for a short-term memory of size ten).

It is interesting to observe that the minimal classification error rate is obtained for size 75 of the short-term memory, not for size 100 as we might have expected. Apparently the reduction of dimensionality from 100 to 75 yields a smaller (simpler) energy landscape where finding a good (if not global) minimum is easier. This more than compensates for the risk of using a frozen long-term memory where reclustering must not be made. However, when the reduction in short-term memory size is carried further the classification error rate increases as expected when fewer and fewer reports take part in simultaneous clustering.

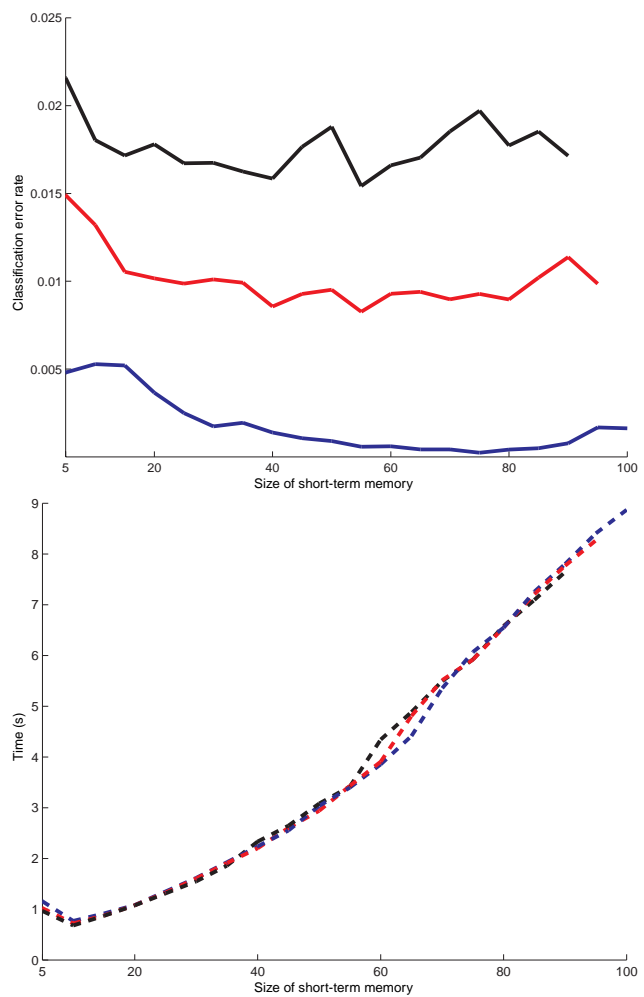

Figure 5. Classification error rate (solid lines) and computation time (dashed lines) for different sizes of short-term memory, when short-term and long-term memories have a joint size of 100 (blue), 95 (red) or 90 (black).

Table 1. Classification error rate and computation time for different configurations of short-term and long-term memories with a joint size of 100.

| Size of short-term memory | Size of long-term memory | Classification error rate (%) | Computation time (s) |
|---|---|---|---|
| 5 | 95 | 0.480 | 1.16 |
| 10 | 90 | 0.528 | 0.769 |
| 15 | 85 | 0.520 | 0.916 |
| 20 | 80 | 0.365 | 1.08 |
| 25 | 75 | 0.250 | 1.35 |
| 30 | 70 | 0.174 | 1.62 |
| 35 | 65 | 0.194 | 1.92 |
| 40 | 60 | 0.139 | 2.24 |
| 45 | 55 | 0.107 | 2.55 |
| 50 | 50 | 0.0900 | 3.04 |
| 55 | 45 | 0.0575 | 3.40 |
| 60 | 40 | 0.0609 | 3.86 |
| 65 | 35 | 0.0426 | 4.42 |
| 70 | 30 | 0.0426 | 5.36 |
| 75 | 25 | 0.0238 | 6.07 |
| 80 | 20 | 0.0422 | 6.55 |
| 85 | 15 | 0.0497 | 7.26 |
| 90 | 10 | 0.0777 | 7.80 |
| 95 | 5 | 0.168 | 8.41 |
| 100 | 0 | 0.163 | 8.87 |

All tests were performed running Allegro CL 5.0.1 using MatLisp 1.0b on a SUN Enterprise 450 computer (480 MHz CPU model Ultrasparc 2, 2.0 GB RAM) with SUN Solaris 8.

# 6  Conclusions

We have demonstrated that an incremental cluster-to-track fusion algorithm based on Dempster-Shafer cluster using a Potts spin neural network (iPDSC) is robust. The robustness is achieved as a second thought on report-to-track association is made for the last reports when new intelligence arrives. A reclustering process with a small classification error rate is used to reconsider the report-to-track association.

# References

[1] G. Shafer, *A Mathematical Theory of Evidence,* Princeton University Press, Princeton, NJ, 1976.

[2] J. Schubert, Cluster-based specification techniques in Dempster-Shafer theory, in *Symbolic and Quantitative Approaches to Reasoning and Uncertainty, Proc. European Conf.* (ECSQARU'95), C. Froidevaux and J. Kohlas (Eds.), Université de Fribourg, Switzerland, 3–5 Jul 1995, pp. 395–404, Springer-Verlag (LNAI 946), Berlin, 1995.

[3] J. Schubert, Fast Dempster-Shafer clustering using a neural network structure, in *Information, Uncertainty and Fusion,* B. Bouchon-Meunier, R.R. Yager and L.A. Zadeh, (Eds.), pp. 419–430, Kluwer Academic Publishers (SECS 516), Boston, MA, 1999.

[4] J. Schubert, Simultaneous Dempster-Shafer clustering and gradual determination of number of clusters using a neural network structure, in *Proc. 1999 Information, Decision and Control Conf.* (IDC'99), Adelaide, Australia, 8–10 Feb 1999, pp. 401–406, IEEE, Piscataway, NJ, 1999.

[5] M. Bengtsson, and J. Schubert, Dempster-Shafer clustering using potts spin mean field theory, *Soft Computing,* Vol. 5, No. 3, pp. 215–228, June 2001.

[6] J. Schubert, Specifying nonspecific evidence, *Int. J. Intell. Syst.,* Vol 11, No. 8, pp. 525–563, Aug 1996.

[7] J. Schubert, Clustering belief functions based on attracting and conflicting metalevel evidence, in *Proc. Ninth Int. Conf. Information Processing and Management of Uncertainty in Knowledge-based Systems* (IPMU 2002), Annecy, France, 1–5 Jul 2002, to appear.

[8] A. Ayoun, and P. Smets, Data association in multi-target detection using the transferable belief model, *Int. J. Intell. Syst.,* Vol. 16, No. 10, pp. 1167–1182, Oct 2001.

[9] P. Smets, Practical uses of belief functions, in *Proc. Fifteenth Conf. Uncertainty in Artificial Intelligence* (UAI'99), K.B. Laskey and H. Prade (Eds.), Stockholm, Sweden, 30 Jul–1 Aug 1999, pp. 612–621, Morgan Kaufmann Publishers, San Francisco, CA, 1999.

[10] P. Smets, Data Fusion in the Transferable Belief Model, in *Proc. Third Int. Conf. Information Fusion* (FUSION 2000), Paris, France, 10–13 Jul 2000, pp. PS/20–33, International Society of Information Fusion, Sunnyvale, CA, 2000.

[11] J. Schubert, Managing inconsistent intelligence, in *Proc. Third Int. Conf. Information Fusion* (FUSION 2000), Paris, France, 10–13 Jul 2000, pp. TuB4/10–16, International Society of Information Fusion, Sunnyvale, CA, 2000.

[12] N. Milisavljevic, Analysis and fusion using belief function theory of multisensor data for close-range humanitarian mine detection, Ph.D. Thesis, ENST 2001 E 012, École Nationale Supérieure des Télécommunications, Paris, 2001.

[13] M. Grabisch, and H. Prade, The correlation problem in sensor fusion in a possibilistic framework, *Int. J. Intell. Syst.*, Vol. 16, No. 11, pp. 1273–1283, Nov 2001.

[14] F.Y. Wu, The Potts model, *Rev. Mod. Phys.,* Vol. 54, No. 1, pp. 235–268, Jan 1982.

[15] J. Schubert, On nonspecific evidence, *Int. J. Intell. Syst.,* Vol 8, No. 6, pp. 711–725, Jul 1993.

[16] C. Peterson, and B. Söderberg, A new method for mapping optimization problems onto neural networks, *Int. J. Neural Syst.,* Vol 1, No. 1, pp. 3–22, 1989.